\documentclass[
]{ceurart}

\usepackage{listings}
\usepackage[inline]{enumitem}

\RedeclareSectionCommand[afterskip=-.5em]{subsubsection}

\usepackage{caption}
\usepackage{subcaption}

\usepackage{xspace}
\newcommand{\clevr}{\texttt{CLEVR}\xspace}

\usepackage{url}
\usepackage{pgfplots}
\pgfplotsset{width=8cm,height=4cm,compat=1.9}

\usepackage{pgf-pie}

\begin{document}

\copyrightyear{2022}
\copyrightclause{Copyright for this paper by its authors.
  Use permitted under Creative Commons License Attribution 4.0
  International (CC BY 4.0).}

\conference{NeSy 2022, 16th International Workshop on Neural-Symbolic Learning and Reasoning, Cumberland Lodge, Windsor, UK}

\title{CLEVR-Math: A Dataset for Compositional Language, Visual and Mathematical Reasoning}


\author[1]{Adam Dahlgren Lindström}[%
orcid=0000-0002-1112-2981,
email=dali@cs.umu.se,
]
\cormark[1]
\address[1]{Umeå University, Sweden}

\author[2]{Savitha Sam Abraham}[%
orcid=0000-0003-3902-2867,
email=savitha.sam-abraham@oru.se,
]
\cormark[1]
\address[2]{Örebro University, Sweden}

\cortext[1]{Corresponding author.}

\begin{abstract}
 We introduce CLEVR-Math, a multi-modal math word problems dataset consisting of simple math word problems involving addition/subtraction, represented partly by a textual description and partly by an image illustrating the scenario. The text describes actions performed on the scene that is depicted in the image. Since the question posed may not be about the scene in the image, but about the state of the scene before or after the actions are applied, the solver envision or imagine the state changes due to these actions. Solving these word problems requires a combination of language, visual and mathematical reasoning. We apply state-of-the-art neural and neuro-symbolic models for visual question answering on CLEVR-Math and empirically evaluate their performances. Our results show how neither method generalise to chains of operations. We discuss the limitations of the two in addressing the task of multi-modal word problem solving. 
\end{abstract}


\maketitle

\section{Introduction}

Math word problems are mathematical problems expressed in natural language. Solving these problems requires one to be able to map the natural language text to a mathematical expression, identifying the known and unknown quantities and the functions to be used. Since this is a task that lies at the intersection of language understanding and reasoning, automatically solving math word problems has been a research topic of interest in Natural Language Processing (NLP) community. Consider the following math word problem, 
\begin{quote}
    \textbf{Problem: }\emph{Adam has three apples, and Eve has five. Eve gives Adam all her apples. How many apples does Adam have, if he \emph{eats} one?}\\
    \textbf{Equation:} $X$ = $3+5-1$
\end{quote}
Minor changes in the text may result in large semantic changes, e.g. changing just one word in the above problem - \emph{eats} to \emph{finds}, will change the equation to $X$ = $3+5+1$. Most of the recent efforts in automatic math word problem solving treat it as a translation task (from word problem to equation) and have employed sequence-to-sequence networks or sequence to tree (generating the expression tree of the equation) networks (\cite{luong2015deep},~\cite{xie2019goal},\cite{zhang2020graph}).   

\begin{wrapfigure}{r}{0.5\textwidth}
\centering
\includegraphics[width=\linewidth]{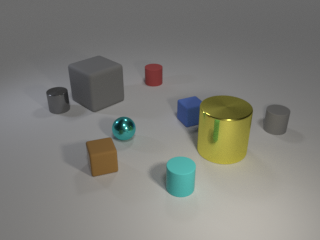}
\caption{CLEVR-Math example question \emph{Take away 2 matte cylinders. How many objects are left?} with corresponding mathematical equation $X = 9 - 2$.}\label{fig1}
\end{wrapfigure}
While text-based math word math problems are a great setting for natural language understanding, it would also be interesting to consider word problems which are accompanied by a diagram, and the information required to derive the solution has to be captured from both its textual and visual representations. That is, part of the problem scenario description is expressed as text and the other part is represented in the form of an image. In this paper, we introduce such a multi-modal math word problem dataset, CLEVR-Math (since it is based on CLEVR dataset~\cite{johnson2017clevr}),  where each problem has a textual and a visual description (image). Figure \ref{fig1} shows an example problem in CLEVR-Math dataset. 


While each instance in CLEVR dataset has an image and a natural language query about the scene depicted in the image, in CLEVR-Math, the natural language query may not be about the scene represented in the image, but about the state of the scene after/before a sequence of actions are applied on the scene. The actions in our case are addition/removal of specific type of objects to/from the original scene. We believe this is an interesting problem setting as the ability to envision changes without them being physically manifested is an important aspect of the human mind. 

Our contributions are two-fold, we
\begin{itemize}
    \item construct an open source multi-modal math word problem dataset, CLEVR-Math and
    \item analyse the performance of state-of-the-art neural and neuro-symbolic (NeSy) solutions for solving such multi-modal problems. 
    \end{itemize}
Our results and analysis shows how both neural and NeSy methods are unable to compositionally generalise to chains of operations.


\section{Related Work}
This section gives an overview of the existing datasets in math word problem solving, multi-modal datasets for visual question answering and existing neural/neuro-symbolic approaches to the tasks that require a combination of visual, language and logical reasoning. 

\subsubsection*{Math Word Problem Solving - Datasets}
Math Word Problem Solving (MAWPS)~\cite{koncel2016mawps} was one of the earlier datasets introduced in the domain and collected around $3320$ single/multi equation word problems involving operators $+$, $-$, $*$, $/$. These word problems were annotated with equations involved and the answer (solution of the equation). More recently, larger datasets like Algebra Question Answering with Rationales (AQuA-RAT)~\cite{ling2017program} were introduced and it has around $100K$ multiple choice questions annotated with equations and a textual explanation for the rationale behind the equations. 
~\cite{patel2021nlp} illustrated the deficiencies in MAWPS dataset by introducing another dataset named  Simple Variations on Arithmetic Math word Problems (SVAMP). SVAMP is created by making minor variations to problems in MAWPS.~\cite{patel2021nlp} showed that state-of-the-art neural solvers trained on MAWPS performs poorly on the SVAMP dataset. 

\subsubsection*{Visual Question Answering (VQA) and Visual Reasoning - Datasets}
One of the first VQA datasets proposed was the DAQUAR dataset~\cite{malinowski2014multi} based on real images of indoor scenes. VQA is another widely used dataset~\cite{antol2015vqa} with images from MS-COCO dataset~\cite{lin2014microsoft}. Questions are manually created and answering these require commonsense knowledge and reasoning. CLEVR dataset~\cite{johnson2017clevr} is based on automatically generated scenes and questions. Such simulated data gives great control over the distribution of instances. One can decide to generate a training set with images having only a specific combination of objects (red cubes and blue cylinders), and a test set with a different combination of objects (red cylinders and blue cubes), as done in, e.g., CLEVR-Hans~\cite{stammer2021right}. This control allows us to study various aspects like compositional generalisation of systems.
Based on these ideas, CLEVR-Math allows us to test the ability of the system to generalise to unseen combinations of actions in the word problem, to e.g. train on single mathematical operations, and test on chains of operations.
Closely related is the CLEVRER (Collision Events for Video Representation and Reasoning) dataset~\cite{yi2019clevrer} and CLEVR-Hyp dataset \cite{sampat2021clevr_hyp}. 
The questions on videos in CLEVRER requires reasoning about the state of objects after an video event, instead of after actions in text as in CLEVR-Math. 
CLEVR-Hyp focus on VQA where reasoning about effects of actions, and CLEVR-Math introduces an additional mathematical reasoning dimension to the problem. 
GQA is another relevant dataset, where real world images are annotated with rich scene graphs and a large set of relations and attributes, and focuses on compositionality in visual reasoning~\cite{hudson2019gqa}.
Graph learning is a heavily studied area, with applications in multimodal domains such as robotics~\cite{9416834,yu2021ernie,wald2020learning,ji2020action}.

Experiments with Kandinsky patterns~\cite{holzinger2019kandinsky} show that neural networks are easily confounded by visual reasoning tasks with shapes, colors, and patterns that can be difficult to distinguish but follow clear rules.
The Winoground dataset~\cite{thrush2022winoground} shows similar results, where no state-of-the-art visual reasoning method is able to distinguish between two confounding captions and images.


\subsubsection*{Existing Approaches to VQA}
Most of the earlier approaches in VQA were based on purely neural models that first encoded the two inputs - the image and the accompanying question into embeddings using networks like Convolutional Neural Networks (CNN) and Long Short Term Memory (LSTM)  networks and then the two embeddings were forwarded to a classifier that would then predict the answer to the question (\cite{ben2017mutan},~\cite{fukui2016multimodal}). Another category of approaches are the attention mechanism-based approaches that identified the regions in the image that were relevant to answering the associated question (\cite{wang2017fvqa},~\cite{shih2016look}). Graph neural networks~\cite{narasimhan2018out} have also been applied in VQA where both text and the image are represented as graphs and a multi-modal vectorial representation is learned that captures the alignment of nodes in the two graphs.~\cite{radford2021learning} introduced the CLIP models where a representation of the image is learned with natural language supervision by leveraging the already available huge datasets for image captioning. More recently, neuro-symbolic approaches have been used in addressing the task of VQA like Neuro Symbolic Concept Learner (NSCL)~\cite{mao2019neuro} and Neuro-Symbolic Visual Question Answering (NS-VQA)~\cite{yi2018neural}. These approaches convert the input image and text into an intermediate semantic representation and then employ a quasi-symbolic program executor to derive an answer from these semantic forms. We use CLIP and NS-VQA as baselines as they are state-of-the-art on multimodal language modelling and on the CLEVR dataset, respectively. 

\section{CLEVR-Math}

We construct the CLEVR-Math dataset as an extension of \clevr by introducing three new functions and 13 templates.
Using the codebase provided with \clevr, we generate new questions based on the original scenes.
We categorise the 13 templates into six types, all based on addition and subtraction.
The domain is restricted to numbers between $0-10$ to conform with \clevr.

\subsubsection*{New \clevr functions:} 
\begin{figure}[h]
    \centering
    \vfill
   \begin{subfigure}[b]{0.42\textwidth}
        \centering
        \includegraphics[width=\linewidth]{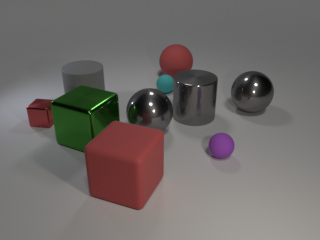}
        \caption{(i) \emph{Remove all gray spheres. How many spheres are there? (3)}, (ii) \emph{Take away 3 cubes. How many objects are there?} (7), (iii) \emph{How many blocks must be removed to get 1 block? (2)}}
        \label{fig:example}
    \end{subfigure}\hfil
    \begin{subfigure}[b]{0.42\textwidth}
        \centering
        \includegraphics[width=\linewidth]{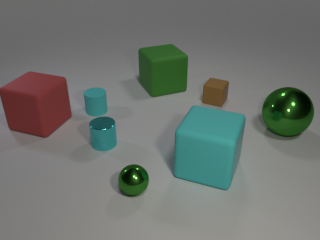}
        \caption{\emph{Take away all large green metallic spheres. Now remove all cyan objects. How many objects are left? (4)}}
        \label{fig:chained}
        \vspace{0.85cm}
    \end{subfigure}
    \caption{Example image-question pairs from CLEVR-Math, \ref{fig:example} showcase addition and subtraction, and \ref{fig:chained} shows multihop reasoning. Answers in parenthesis.}
\end{figure}

The three functions that we implement are - \texttt{subtraction} and \texttt{addition} to perform subtraction and addition, and \texttt{choose} to operate on subsets of objects.
Instead of removing all blue spheres, \texttt{choose} allows us to remove a random number of a specific type of object, e.g. 2 blue spheres out of 4.
The random number generated by \texttt{choose} replaces a questions ``X" placeholder during generation. 
Figure~\ref{fig:example} shows three examples of subtraction, and Figure~\ref{fig:chained} shows a question requiring multihop reasoning.
Appendix~\ref{sec:samples} includes more samples from the test set.

\subsubsection*{Question Categories:} The different question categories are shown in Table \ref{tab:question:types}. 
\begin{itemize}
    \item \textit{\textbf{Remove group}}: All objects belonging to a specific group are removed from the scene.  
    \item \textit{\textbf{Insertion}}: A specific number of objects are added to the scene. 
    \item \textit{\textbf{Count backwards}}: The query is about the change - that is the number of objects added/removed from the scene to get a goal state. 
    \item \textit{\textbf{Remove subset}}: A specific number of objects are removed from the scene.
    \item \textit{\textbf{Adversarial questions}}: These are trick questions where the actions may be performed on one object, but the query is about an object that is not affected by the action. The adversarial actions are always on objects that are seen in the image.
    \item \textit{\textbf{Multi-hop}}: In contrast to the above questions, multi-hop questions perform sequences of actions (insertion, removal) on the objects. Such questions with chained functions help us test a model's ability to generalise to infinite combinations of operations.
\end{itemize}

\begin{table}[h]
    \centering
    \begin{tabular}{l|l}
         \textbf{Type} & \textbf{Templates} \\
         \hline
        \textbf{Remove group} &\\
        &"Remove all <C> <S>s. How many <S>s are there?"\\
        &"Take away all <Z> <C> <M> <S>s. How many <S>s are there?" \\
        &"Take away X <C> <S>s. How many objects are there?" \\
        &"Take away all <C> <S>s. How many objects are there?" \\

        \hline
        \textbf{Insertion} & \\
        &"Add X <Z> <C> <M> <S>s. How many <Z> <C> <M> <S>s are here?" \\
        &"Add X <Z> <C> <M> <S>s. How many objects are there?" \\

        \hline            
        \textbf{Count backwards} & \\
        &"How many <C> <S>s must be removed to get X <C> <S>s?" \\
        &"Take away <C> <S>s. How many were removed if there are X <C> <S>s left?" \\
            
        \hline
        \textbf{Multi-hop} & \\
        &"Take away all <Z> <C> <M> <S>s. Remove all \\
        &<Z2> <C2> <M2> <S2>s. How many objects are left?" \\

        \hline            
        \textbf{Remove subset} &\\
        &"Remove X <S>s. How many <S>s are there?" \\
            
        \hline
        \textbf{Adversarial questions} & \\
        &"Remove all <C1> <S1>s. Remove all <C2> <S2>s. How many <S1>s are left?" \\
        &"Remove all <C1> <S1>s. How many <C2> <S2>s are left?" \\
    \end{tabular}
    \caption{An overview of the different templates implemented by CLEVR-Math. <Z>, <C>, <M>, <S> are instantiated to size, color, material, and shape during the question generation.}
    \label{tab:question:types}
\end{table}
Each problem in the dataset is also annotated with it's equivalent functional program based on the CLEVR functions described in the previous section. For example, consider the question from insertion category and it's program (the arguments of an instruction refer to another instruction - indicating it's input is the output of the referred instruction):
\begin{quote}
\textbf{Q}: \textit{Add 3 blue cylinders. How many cylinders are there?} \\
\textbf{Program}: \textit{1. scene, 2. choose[3], 3. count(1), 4. filter\_cylinder(1), 5. count(4), addition(2, 5)} 
\end{quote}
The program contains the \texttt{choose} function -  \texttt{choose[$i$]} operator returns $i$ ($i=3$ in this case).

\subsubsection*{Question generation.} 
\begin{figure}[h]
    \centering
    \begin{subfigure}[b]{0.7\textwidth}
        \centering
        \begin{tikzpicture}
            \begin{axis} [ybar,ymin=0,xmin=-0.5, xmax=4, 
            xmajorticks=false,
            width=9cm, height=4.5cm,
            x tick label style={rotate=90},
            ]
            \addplot coordinates {
                (0.4, 188437) 
                (0.62, 189250)
                (0.84, 189039)
                };
            \addplot coordinates {
                (1.0, 105826)
                (1.22, 106086)
                
            };
            \addplot coordinates {
                (1.4, 107817)
                (1.62, 103333)
            };
            \addplot coordinates {
                (1.8, 61115)
                (2.02, 60864)
                (2.22, 61010)
                (2.42, 61635)
                (2.62, 61255)
                (2.82, 61788)
                (3.02, 61373)
                (3.22, 61029)
            };
            \legend {Shape, Material, Size, Color}
            \end{axis}
            
            
            
        \end{tikzpicture}

        \caption{Attribute distribution per category, showing even allocations.}
        \label{fig:stats:attributes}
    \end{subfigure}
    \hfill
    \begin{subfigure}[b]{0.4\textwidth}
        \centering
        \begin{tikzpicture}
            \begin{axis} [ybar, ymin=0, xmin=-1, xmax=11, width=6cm, height=3.5cm]
            \addplot coordinates {
                (0,17050)
                (1, 55647)
                (2, 78468)
                (3, 82180)
                (4, 77339)
                (5, 62817)
                (6, 50569)
                (7, 40674)
                (8, 36053)
                (9, 31043)
                (10, 24242)
                };
            \end{axis}
            
            
            
        \end{tikzpicture}

        \caption{Answer distribution, from 0 to 10.}
        \label{fig:stats:answers}
    \end{subfigure}
      \begin{subfigure}[b]{0.4\textwidth}
    \centering
            \begin{tikzpicture}
            \begin{axis} [ybar, width=6cm, height=3.5cm
            ]
            \addplot coordinates { 
                (7,3638)
                (8,112733)
                (9,154008)
                (10, 76438)
                (11, 36041)
                (12, 33895)
                (13,66136)
                (14, 30357)
                (15, 24710)              
                (16, 14625)
                (17, 3501) 
                };
            \end{axis}
        \end{tikzpicture}
            
            
        \caption{Distribution of number of words.}
        \label{fig:narrow:lang}
    \end{subfigure}
    \begin{subfigure}[b]{0.7\textwidth}
    \centering
            \begin{tikzpicture}
            \begin{axis} [ybar, xmin=0, xmax=4.5, xticklabels={}, xmajorticks=false, width=11cm, height=3.8cm]
            \addplot coordinates {
                (0.4, 31568) 
                (0.62, 31717)
                (0.82, 64430)
                (1.02, 30788)
                (1.22, 3175)
                };
            \addplot coordinates {
                (1.4, 67826)
                (1.62, 67754)
                (1.82, 58061)
                
            };
            \addplot coordinates {
                (2.0, 40455)
                (2.22, 24445)
                (2.42, 280)
            };
            \addplot coordinates {
                (2.6, 67897)
            };
            \legend {Subtraction, Addition, Adversarial, Multihop}
            \end{axis}
            
        \end{tikzpicture}
        \caption{Template distribution over categories of templates. Each bar corresponds to a template in each respective category. We see that subset subtraction (i.e., \emph{remove 2 blue cubes}) is underrepresented.}
        \label{fig:stats:templats}
    \end{subfigure}
   
           



                
           
            

    \caption{The attributes are used evenly throughout the dataset, whereas the answers are biased towards the smaller numbers. The numbers are aggregated over all splits. }
    \label{fig:stats}
\end{figure}
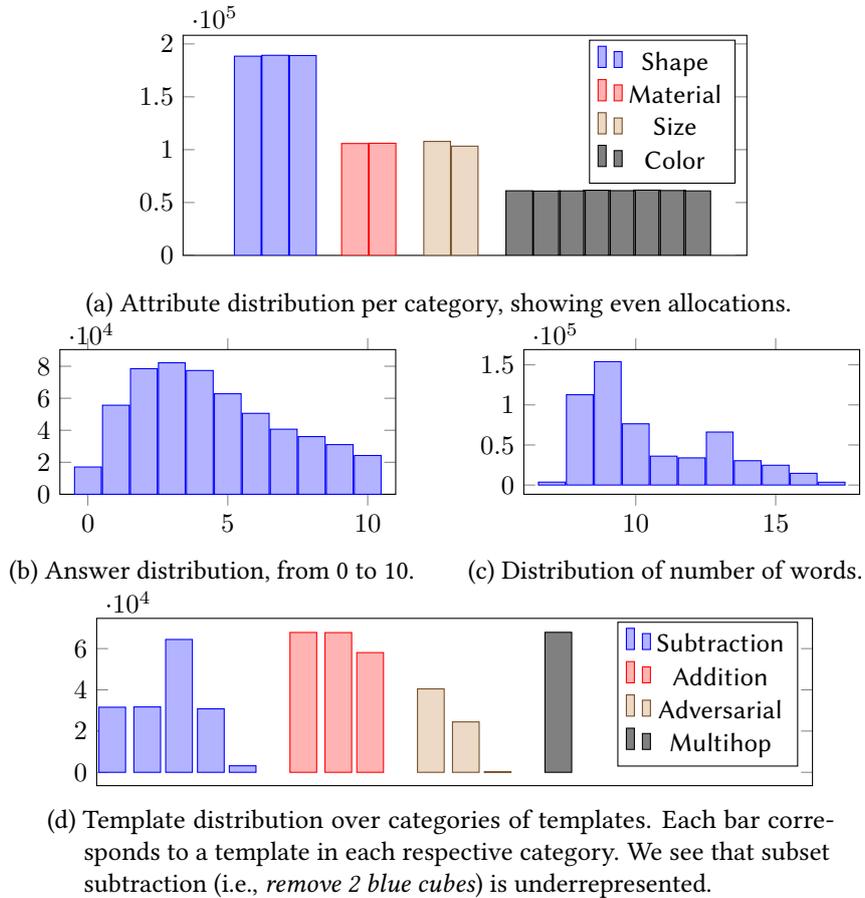

To support greater linguistic variation, we add synonyms for addition and subtraction to the template engine. \emph{Subtract} can be replaced with \emph{remove, take away} and \emph{withdraw}, and \emph{addition} with \emph{introduce, } and \emph{insert}.
We use the same training and validation scenes as \clevr, and generate 5000 new scenes as test data.
Figure~\ref{fig:stats} show the distribution of attributes, words, templates and answers in CLEVR-Math, aggregated over the training, validation, and test data.
The distribution is reflected in each of the splits.

        

There are 50 words in the CLEVR-Math vocabulary, where the narrow language puts focus on the mathematical reasoning rather than advanced language capabilities.
Figure~\ref{fig:narrow:lang} show that most questions are 8-9 words long, with a second peak at 13 for the multihop questions.

\begin{table}[h]
    \centering
    \begin{tabular}{c|c|c|c}
        \textbf{Template} & \textbf{Train} & \textbf{Validation} & \textbf{Test}\\
        \hline
         Subtraction & 229364 & 49149 & 3281\\
         Addition &193641 & 41600 & 2752\\
         Adversarial & 65180 & 13900 & 950\\
         Multihop & 67897 & 14553 & 972 \\
    \end{tabular}
    \caption{Distribution of templates in each data split.}
    \label{tab:template:distribution}
\end{table}
Table~\ref{tab:template:distribution} shows the distribution of templates, with an approximately equal amount of questions for subtraction and addition, and similarly for adversarial and multihop questions.
The ratios are consistent between splits.
To test multihop reasoning and compositional generalisability we generate train-validation-test with only singlehop questions in training and validation, and only multihop questions in the test data.
Thus, a model using the \texttt{CLEVR-Math-multihop} configuration must solve the multihop questions in a zero-shot fashion.

\subsubsection*{Open sourcing data.} We open source CLEVR-Math as a Huggingface dataset~\footnote{\url{https://huggingface.co/datasets/dali-does/clevr-math}} with two configurations; \texttt{CLEVR-Math} and \texttt{CLEVR-Math-multihop}.
The extended \clevr source code is available on Github~\footnote{\url{https://github.com/dali-does/clevr-math}}.
Table~\ref{tab:data:huggingface} shows the Huggingface dataset card for CLEVR-Math.
The template feature allows for filtering to perform, e.g., only singlehop training and multihop testing.
\begin{table}[h]
    \centering
    \begin{tabular}{c|c|l}
        \textbf{Feature} & \textbf{Type} & \textbf{Example}\\
        \hline
        template & String & \texttt{subtraction-multihop} \\
        id & String & \texttt{CLEVR\_math\_test\_000010.png}\\
        question & String & \emph{Remove 5 spheres. How many objects are there?}\\
        image & image path & \url{CLEVR_v1.0/images/train/CLEVR_new_000010.png}\\
        label & int64, 0-10 & 5\\
    \end{tabular}
    \caption{Huggingface dataset card for CLEVR-Math.}
    \label{tab:data:huggingface}
\end{table}





\section{Experiments}

CLIP~\cite{radford2021learning} is used as a neural baseline. Questions and images are embedded using CLIP, and an additional classification layer is added to predict the correct answer. Fine tuning CLIP on CLEVR-Math as a masked language task before adding classification gave no significant improvements, while consuming significantly more computational resources. CLIP and this classification layer is trained jointly for $10$ epochs with early stopping using a batch size of $64$.

NS-VQA~\cite{yi2018neural} is used as the neuro-symbolic baseline. Here, a mask-RCNN~\cite{he2017mask} is trained independently to convert an image to a scene graph. 
In our experiments, we skip this step and use the actual scene graphs associated with images. The question is parsed into a functional program by a sequence to sequence (Seq2Seq) network based on Bi-LSTM. A quasi-symbolic program executor executes the program generated on the scene graph of the image to return an answer. The Seq2Seq network is pre-trained in a fully supervised fashion by providing it a few examples (around 60 examples) of $(question,$ $program)$ pairs. The pre-trained network is then trained further using REINFORCE algorithm that returns a reward based on whether the program generated could derive the expected answer or not. Supervised pretraining and REINFORCE were run for $1000$ and $5000$ iterations, respectively, with a batch size of $128$.
Both CLIP and NS-VQA models were trained on a NVIDIA Tesla P100 GPU computing processor.

Each model is evaluated on each question category, and are trained on 2500, 5000, 10000, and 20000 samples to see the influence of the amount of data.
For multihop, training and validation sets with and without multihop questions are used, with the latter named \emph{multihop (0-shot)}.

\subsection{Results}
Table~\ref{tab:baseline:templates} shows the accuracy of CLIP and NS-VQA on the different categories as well as an aggregated accuracy over the entire dataset. 
Both the models were trained on $10,000$ samples. 
NS-VQA performs better than CLIP models for most templates apart from multihop. NS-VQA performs better on subtraction and adversarial problems (both based on `subtraction' CLEVR function) than addition problems. This could be because the functional programs for addition problems always contain a \texttt{choose} operator. It is important to identify the argument to \texttt{choose} operator from the problem statement (which is mostly one of the numerical quantities in the word problem) to arrive at the correct answer. Unlike this, there are subtraction and adversarial problems (in remove group) that do not have a \texttt{choose} operator in the program.
\begin{table}[h]
    \centering
    \begin{tabular}{c|c|c|c|c|c|c}
        \textbf{Model} & \textbf{All} & \textbf{Addition} & \textbf{Subtraction} & \textbf{Adversarial} & \textbf{Multihop} & \textbf{Multihop (0-shot)}\\
        \hline
        NS-VQA & 0.8840  & 0.9781 & 0.9948 &  0.9957  & 0.286 &  0.267\\
        CLIP &  0.3464 & 0.5699 & 0.3019 &  0.2848  & 0.272 & 0.238\\
    \end{tabular}
    \caption{Accuracy on the CLEVR-Math dataset, shown for each template group and aggregated over all templates.}
    \label{tab:baseline:templates}
\end{table} 
Neither of the methods perform well on the multi-hop questions, with a clear degradation in the performance for NS-VQA. This is because the question parser of NS-VQA relies on a Seq2Seq network that does not generalize compositionally~\cite{lake2018generalization}.
\clevr focus on visual attribute compositionality, and the multihop reasoning introduces higher demands on linguistic compositionality.
When multihop questions are included in the training and validation data, naturally both methods improve their performance.

To gain further insight into CLIPs' performance on CLEVR-Math, Appendix~\ref{sec:conf} shows a confusion matrix from training CLIP on 20 000 samples and evaluating on all question categories.
These results show that most errors made by CLIP is off by ones.
This reflects the generative nature of such models, in how they can get the context correct but sometimes miss out on details.
We also see how CLIP focus on learning in the range 1-5, reflecting that these problems represent a majority of the problems.

\begin{table}[h]
    \centering
    \begin{tabular}{c|c|c|c|c}
        \textbf{Model} & \textbf{2500}  & \textbf{5000} & \textbf{10000} & \textbf{20000}\\
        \hline
        NS-VQA & 0.6283 & 0.8840 & 0.6795  & 0.6118\\
        CLIP & 0.2918 & 0.3184 & 0.3528 & 0.3464\\
    \end{tabular}
    \caption{Accuracy over all templates for different dataset sizes.}
    \label{tab:baseline:sizes}
\end{table}
Table~\ref{tab:baseline:sizes} shows how different training sizes influence the accuracy. We can see that NS-VQA achieves high accuracy from relatively few examples and plateaus, which is consistent with the original results on \clevr.
It also seems like NS-VQA is overfitting with more data given, and one hypothesis is that more emphasise is put on the program, but that they are similar enough to confound NS-VQA.
In \clevr, the different questions were more distinguishable from a program perspective.
CLIP scales with the number of samples, but plateaus at a much lower accuracy.
We note that a larger number of samples could lead to similar performance for CLIP, but at the cost of more computational resources. 

\begin{figure}[htb]
    \centering 
\begin{subfigure}{0.4\textwidth}
  \includegraphics[width=\linewidth]{figures/samples/incorrect/CLEVR_test_000270.png}
  \caption{\textit{Subtract all small purple matte blocks. Subtract all blocks. How many objects are left?} was answered by CLIP with 3 instead of 2.}
  \label{fig:multihop:wrong:1}
\end{subfigure}\hfil 
\begin{subfigure}{0.4\textwidth}
  \includegraphics[width=\linewidth]{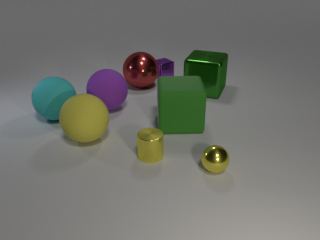}
  \caption{\textit{Subtract all red metallic objects. Subtract all yellow objects. How many objects are left?} was answered with 9 instead of 5 by NS-VQA.}
  \label{fig:2}
\end{subfigure}
\caption{Examples of when CLIP and NS-VQA fails on multihop questions.}\label{fig:multihop:wrong}
\end{figure}
We randomly sample 20 correct and 20 incorrect answers from the multihop test data for both CLIP and NS-VQA.
Appendix~\ref{app:multihop} contains a subset of those samples, and Figure~\ref{fig:multihop:wrong} illustrates two incorrect answers.
There are no clear patterns of failures, such as only performing one of the actions, but we notice multiple instances where CLIP fails to perform overlapping subtraction, or subtraction when no objects match the description.
Another observation is that half of the 20 incorrect answers from CLIP, where on images with only three objects.
Scenes with few objects have a much smaller possible action space associated to it, meaning that there is less room for error.
In Figure~\ref{fig:multihop:wrong:1}, there are no purple matte blocks to remove, so the corresponding equation is $3-0-1=2$.

\section{Conclusions and Future Work}
We introduced a new dataset, CLEVR-Math, containing word math problems about visual scenes. 
Our results show that the state-of-the-art NeSy model, NS-VQA, achieves higher accuracy on CLEVR-Math with less data and computational resources, than the neural model, CLIP.
This is further evidence that neural methods, such as CLIP, are lacking in reasoning capabilities, even after fine tuning.
Given that NS-VQA uses perfect scene graphs, the comparison is not completely fair.
We still expect the results of learning end-to-end to be consistent with the current results in alignment with the original results on CLEVR for NS-VQA.

CLEVR-Math successfully introduces a focused benchmark for learning and reasoning in multimodal data.
There are a few natural extensions to this work, both on further development of the dataset and on evaluation.
Extending the benchmark to answers outside of the range 0-10 would provide a more challenging domain, and providing scene graphs for each step of the reasoning chain could open up for other methods.
The empirical results show that neither of the models could generalize to chained actions. Hence, it is also of research interest to design neuro-symbolic models where language perception is tackled in a more generalizable manner. 
Focus should lie on the representations (symbols) that are learned. 
Other interesting directions is to introduce a representation that is manipulated internally according to the actions as they are read.
Adding longer chains of operations, or chains with alternating subtraction and addition, would put even more emphasise on the reasoning capabilities.
Finally, there is an opportunity to add confounding information to test the robustness, e.g. by associating each shape with a fixed color during training and randomise it during testing.


\section{Acknowledgements}
This work was partially supported by the Wallenberg AI, Autonomous Systems and Software Program (WASP) funded by the Knut and Alice Wallenberg Foundation. 

\bibliography{clevr-math}

\clearpage

\appendix

\section{Examples}\label{sec:samples}

\begin{figure}[htb]
    \centering 
\begin{subfigure}{0.25\textwidth}
  \includegraphics[width=\linewidth]{figures/CLEVR_test_000000}
  \caption{Subtract all gray cylinders. Subtract all gray cubes. How many cylinders are left? (3)}
  \label{fig:1}
\end{subfigure}\hfil 
\begin{subfigure}{0.25\textwidth}
  \includegraphics[width=\linewidth]{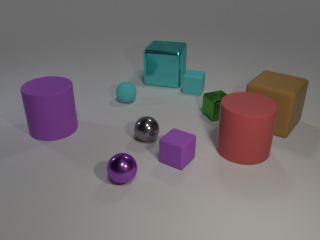}
  \caption{How many cyan cubes must be subtracted to get 1 cyan cubes? (1)\\}
  \label{fig:2}
\end{subfigure}\hfil 
\begin{subfigure}{0.25\textwidth}
  \includegraphics[width=\linewidth]{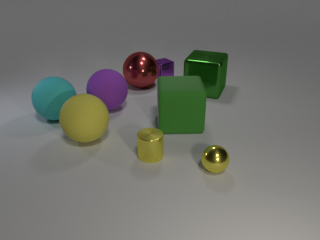}
  \caption{Subtract all purple cylinders. Subtract all yellow blocks. How many cylinders are left? (1)}
  \label{app:fig:examples}
\end{subfigure}

\medskip
\begin{subfigure}{0.25\textwidth}
  \includegraphics[width=\linewidth]{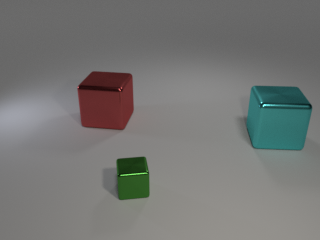}
  \caption{Add 2 large cubes. How many objects exist? (5)}
  \label{fig:4}
\end{subfigure}\hfil 
\begin{subfigure}{0.25\textwidth}
  \includegraphics[width=\linewidth]{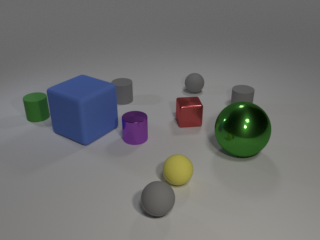}
  \caption{Subtract 4 cylinders. How many cylinders are left? (0)}
  \label{fig:5}
\end{subfigure}\hfil 
\begin{subfigure}{0.25\textwidth}
  \includegraphics[width=\linewidth]{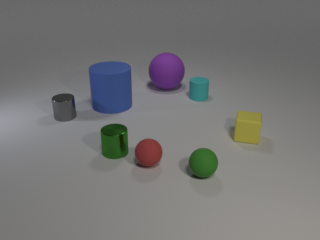}
  \caption{Subtract 0 brown blocks. How many objects are left (8)}
  \label{fig:6}
\end{subfigure}
\caption{Samples from the test set of CLEVR-Math.}
\label{fig:test:samples}
\end{figure}

Figure~\ref{app:fig:examples} show some examples of questions from different templates in CLEVR-Math.
\clearpage

\section{Incorrect answers on multihop questions}\label{app:multihop}

\subsection{CLIP}
\begin{figure}[htb]
    \centering 
\begin{subfigure}{0.25\textwidth}
  \includegraphics[width=\linewidth]{figures/samples/incorrect/CLEVR_test_000270.png}
  \caption{\tiny\textit{Subtract all small purple matte blocks. Subtract all blocks. How many objects are left?} was incorrectly answered with 3 instead of 2 by CLIP.}
  \label{samples:incorrect:1}
\end{subfigure}\hfil 
\begin{subfigure}{0.25\textwidth}
  \includegraphics[width=\linewidth]{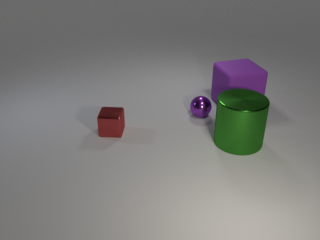}
  \caption{\tiny\textit{Subtract all tiny balls. Subtract all big matte blocks. How many objects are left?} was incorrectly answered with 3 instead of 2 by CLIP.}
  \label{fig:2}
\end{subfigure}\hfil 
\begin{subfigure}{0.25\textwidth}
  \includegraphics[width=\linewidth]{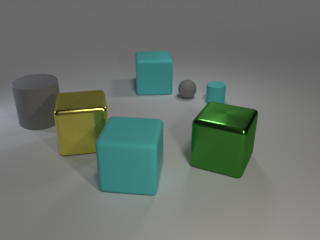}
  \caption{\tiny\textit{Subtract all spheres. Subtract all big green shiny cubes. How many objects are left?} was incorrectly answered with 4 instead of 5 by CLIP.}
  \label{fig:3}
\end{subfigure}

\medskip
\begin{subfigure}{0.25\textwidth}
  \includegraphics[width=\linewidth]{figures/samples/incorrect/CLEVR_test_000295.png}
  \caption{\tiny\textit{Subtract all tiny shiny balls. Subtract all purple objects. How many objects are left?} was incorrectly answered with 4 instead of 8 by CLIP.}
  \label{samples:incorrect:1}
\end{subfigure}\hfil 
\begin{subfigure}{0.25\textwidth}
  \includegraphics[width=\linewidth]{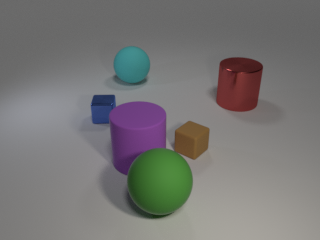}
  \caption{\tiny\textit{Subtract all tiny rubber blocks. Subtract all big cyan rubber things. How many objects are left?} was incorrectly answered with 5 instead of 4 by CLIP.}
  \label{samples:incorrect:1}
\end{subfigure}\hfil 
\begin{subfigure}{0.25\textwidth}
  \includegraphics[width=\linewidth]{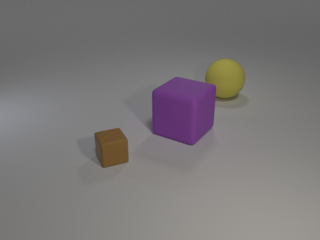}
  \caption{\tiny\textit{Subtract all gray balls. Subtract all small cylinders. How many objects are left?} was incorrectly answered with 3 instead of 2 by CLIP.}
  \label{samples:incorrect:1}
\end{subfigure}

\medskip
\begin{subfigure}{0.25\textwidth}
  \includegraphics[width=\linewidth]{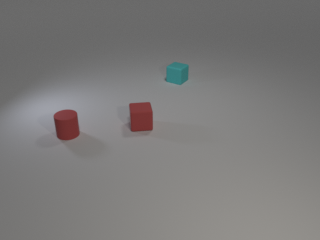}
  \caption{\tiny\textit{Subtract all gray balls. Subtract all small cylinders. How many objects are left?} was incorrectly answered with 3 instead of 2 by CLIP.}
  \label{samples:incorrect:1}
\end{subfigure}\hfil 
\begin{subfigure}{0.25\textwidth}
  \includegraphics[width=\linewidth]{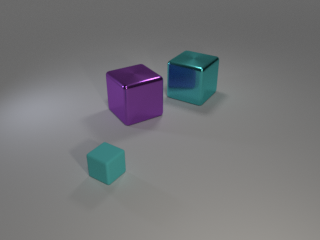}
  \caption{\tiny\textit{Subtract all large cyan objects. Subtract all small rubber blocks. How many objects are left?} was incorrectly answered with 3 instead of 1 by CLIP.}
  \label{samples:incorrect:1}
\end{subfigure}\hfil 
\begin{subfigure}{0.25\textwidth}
  \includegraphics[width=\linewidth]{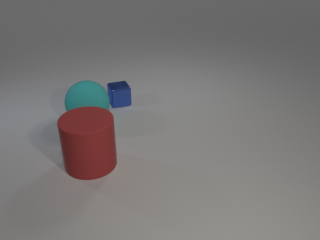}
  \caption{\tiny\textit{Subtract all spheres. Subtract all cyan things. How many objects are left?} was incorrectly answered with 3 instead of 2 by CLIP.}
  \label{samples:incorrect:1}
\end{subfigure}

\medskip
\begin{subfigure}{0.25\textwidth}
  \includegraphics[width=\linewidth]{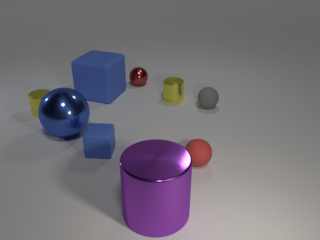}
  \caption{\tiny\textit{Subtract all big green cylinders. Subtract all big blue cubes. How many objects are left?} was incorrectly answered with 5 instead of 8 by CLIP.}
  \label{samples:incorrect:1}
\end{subfigure}\hfil 
\begin{subfigure}{0.25\textwidth}
  \includegraphics[width=\linewidth]{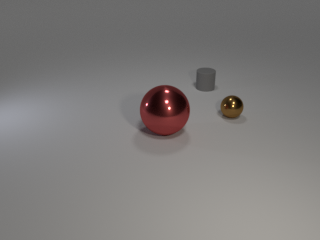}
  \caption{\tiny\textit{Subtract all brown things. Subtract all large red spheres. How many objects are left?} was incorrectly answered with 3 instead of 1 by CLIP.}
  \label{samples:incorrect:1}
\end{subfigure}\hfil 
\begin{subfigure}{0.25\textwidth}
  \includegraphics[width=\linewidth]{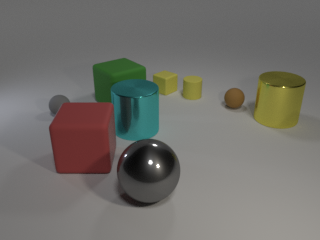}
  \caption{\tiny\textit{Subtract all gray objects. Subtract all big green objects. How many objects are left?} was incorrectly answered with 4 instead of 6 by CLIP.}
  \label{samples:incorrect:1}
\end{subfigure}

\caption{Sampling of incorrect answers by CLIP on multihop.}\label{app:multihop:clip}
\end{figure}

Figure~\ref{app:multihop:clip} shows random samples of when CLIP fails to answer multihop questions correctly.

\clearpage

\subsection{NS-VQA}
\begin{figure}[htb]
    \centering 
\begin{subfigure}{0.25\textwidth}
  \includegraphics[width=\linewidth]{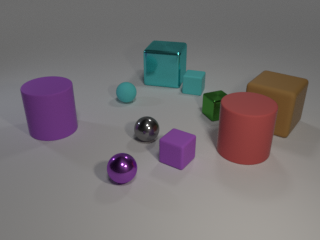}
  \caption{\tiny\textit{Subtract all large purple objects. Remove all green metallic objects. How many objects are left?} was incorrectly answered with 7 instead of 8 by NS-VQA.}
  \label{samples:incorrect:1}
\end{subfigure}\hfil 
\begin{subfigure}{0.25\textwidth}
  \includegraphics[width=\linewidth]{figures/samples/ns-vqa/incorrect/CLEVR_test_000002.png}
  \caption{\tiny\textit{Subtract all red metallic objects. Subtract all yellow objects. How many objects are left?} was incorrectly answered with 9 instead of 5 by NS-VQA.}
  \label{fig:2}
\end{subfigure}\hfil 
\begin{subfigure}{0.25\textwidth}
  \includegraphics[width=\linewidth]{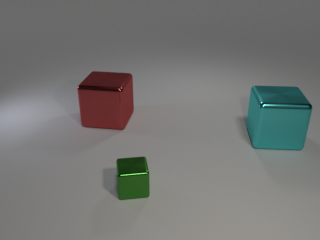}
  \caption{\tiny\textit{Subtract all tiny green metallic cubes. Subtract all large brown blocks. How many objects are left?} was incorrectly answered with 0 instead of 2 by NS-VQA.}
  \label{fig:3}
\end{subfigure}

\medskip
\begin{subfigure}{0.25\textwidth}
  \includegraphics[width=\linewidth]{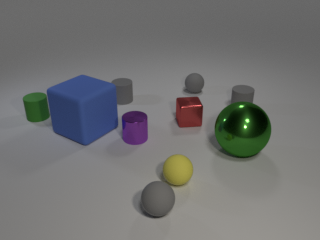}
  \caption{\tiny\textit{Subtract all small red objects. Subtract all tiny metal cylinders. How many objects are left?} was incorrectly answered with 6 instead of 8 by NS-VQA.}
  \label{samples:incorrect:1}
\end{subfigure}\hfil 
\begin{subfigure}{0.25\textwidth}
  \includegraphics[width=\linewidth]{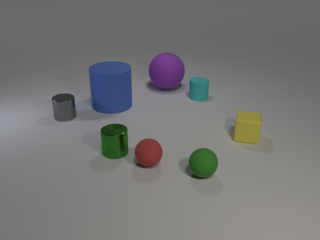}
  \caption{\tiny\textit{Subtract all cylinders. Subtract all purple objects. How many objects are left'?} was incorrectly answered with 7 instead of 3 by NS-VQA.}
  \label{samples:incorrect:1}
\end{subfigure}\hfil 
\begin{subfigure}{0.25\textwidth}
  \includegraphics[width=\linewidth]{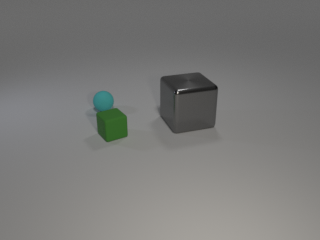}
  \caption{\tiny\textit{Subtract all blue metal cylinders. Subtract all gray objects. How many objects are left?} was incorrectly answered with 1 instead of 2 by NS-VQA.}
  \label{samples:incorrect:1}
\end{subfigure}

\medskip
\begin{subfigure}{0.25\textwidth}
  \includegraphics[width=\linewidth]{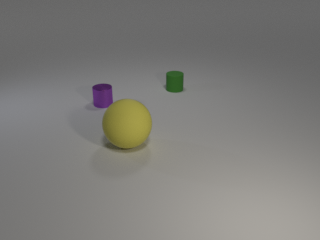}
  \caption{\tiny\textit{Subtract all large rubber spheres. Subtract all blue blocks. How many objects are left?} was incorrectly answered with 3 instead of 2 by NS-VQA.}
  \label{samples:incorrect:1}
\end{subfigure}\hfil 
\begin{subfigure}{0.25\textwidth}
  \includegraphics[width=\linewidth]{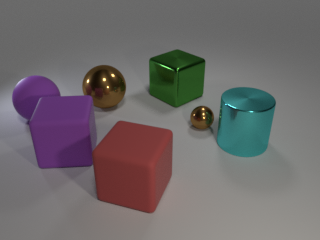}
  \caption{\tiny\textit{Subtract all big gray blocks. Subtract all large cylinders. How many objects are left?} was incorrectly answered with 4 instead of 6 by NS-VQA.}
  \label{samples:incorrect:1}
\end{subfigure}\hfil 
\begin{subfigure}{0.25\textwidth}
  \includegraphics[width=\linewidth]{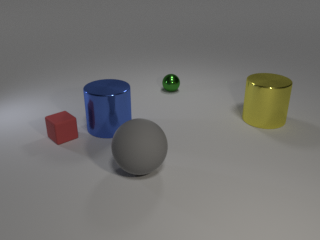}
  \caption{\tiny\textit{Subtract all tiny blocks. Subtract all small balls. How many objects are left?} was incorrectly answered with 4 instead of 3 by NS-VQA.}
  \label{samples:incorrect:1}
\end{subfigure}

\medskip
\begin{subfigure}{0.25\textwidth}
  \includegraphics[width=\linewidth]{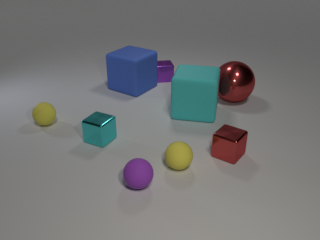}
  \caption{\tiny\textit{Subtract all small purple blocks. Subtract all matte objects. How many objects are left?} was incorrectly answered with 7 instead of 3 by NS-VQA.}
  \label{samples:incorrect:1}
\end{subfigure}\hfil 
\begin{subfigure}{0.25\textwidth}
  \includegraphics[width=\linewidth]{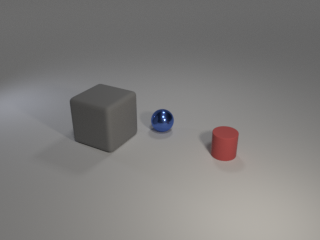}
  \caption{\tiny\textit{Subtract all tiny red rubber objects. Subtract all small blue balls. How many objects are left?} was incorrectly answered with 2 instead of 1 by NS-VQA.}
  \label{samples:incorrect:1}
\end{subfigure}\hfil 
\begin{subfigure}{0.25\textwidth}
  \includegraphics[width=\linewidth]{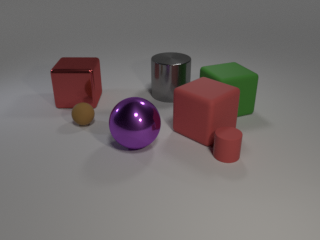}
  \caption{\tiny\textit{Subtract all cyan cylinders. Subtract all small rubber objects. How many objects are left?} was incorrectly answered with 6 instead of 5 by NS-VQA.}
  \label{samples:incorrect:1}
\end{subfigure}

\caption{Sampling of incorrect answers by NS-VQA on multihop.}\label{app:multihop:nsvqa}
\end{figure}

Figure~\ref{app:multihop:nsvqa} shows random samples of when NS-VQA fails to answer multihop questions correctly.
\clearpage

\section{CLIP confusion matrix}\label{sec:conf}

\begin{figure}[h]
    \centering
    \includegraphics[width=\textwidth]{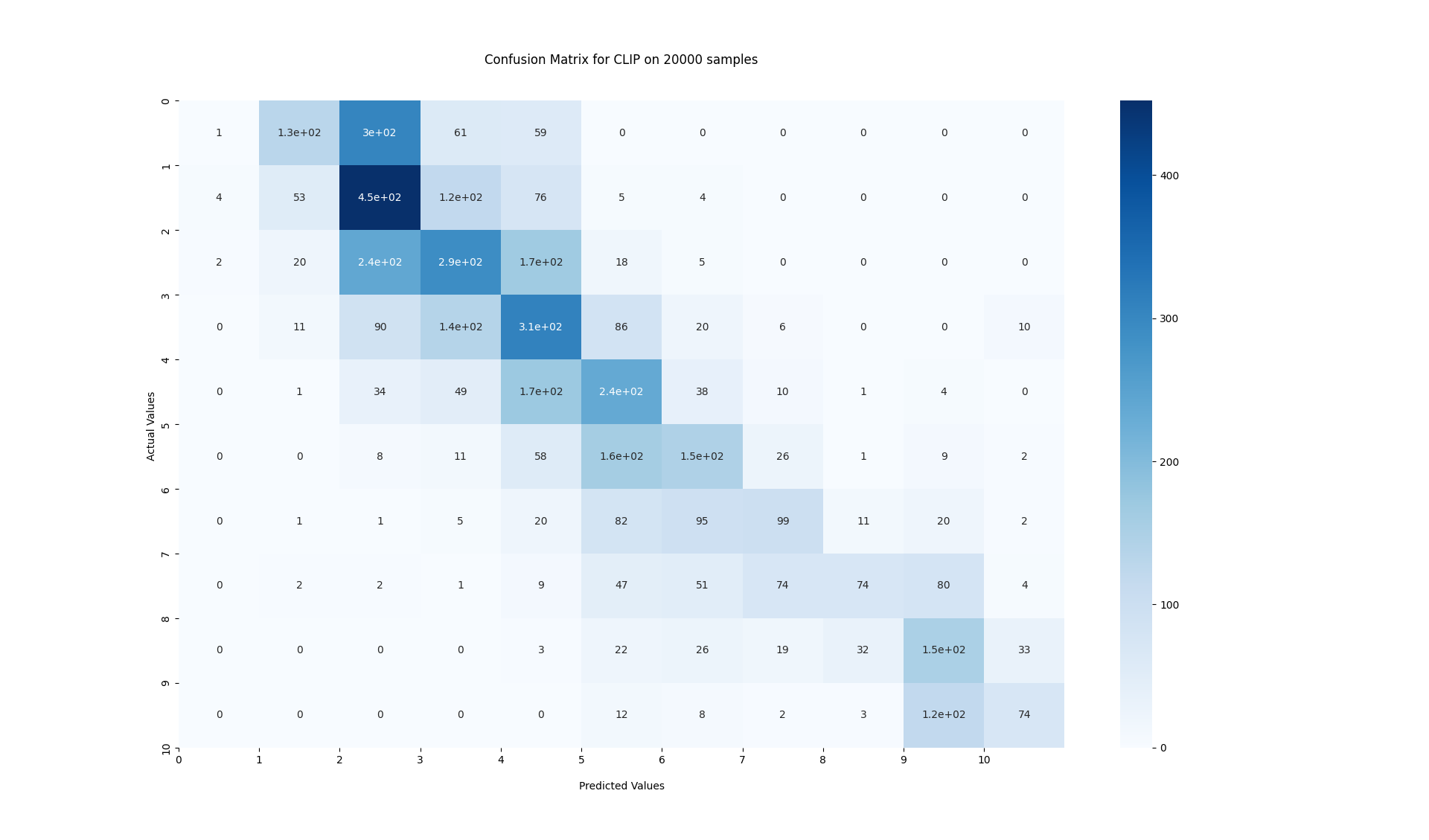}
    \caption{Confusion matrix for CLIP trained on 20 000 samples.}
    \label{fig:conf:matrix}
\end{figure}

Figure~\ref{fig:conf:matrix} shows a confusion matrix indicating that CLIP is learning something for all labels.
It also shows that when an answer is wrong, it is off by one.
The confusion matrix also reflects the distribution over answers, showing that most answers are considered by CLIP to lie in the range 1-5.

\end{document}